\documentclass{article}
\usepackage[utf8]{inputenc}
\usepackage{graphicx}
\usepackage{subcaption}
\makeatletter

\title{Predicting Popularity of Images Over 30 Days}
\author{Amartya Dutta\\
Undergraduate Student\\ Department of Computer Science and Engineering\\Indian Institute of Information Technology Guwahati\\\and
Dr. Ferdous Ahmed Barbhuiya\\
Associate Professor\\Department of Computer Science and Engineering\\Indian Institute of Information Technology Guwahati\\
}
\date{}

\begin{document}

\maketitle 
\begin{abstract}
    The current work deals with the problem of attempting to predict the popularity of images before even being uploaded. This method is specifically focused on Flickr images. Social features of each image as well as that of the user who had uploaded it, have been recorded. The dataset also includes the engagement score of each image which is the ground truth value of the views obtained by each image over a period of 30 days. The work aims to predict the popularity of images on Flickr over a period of 30 days using the social features of the user and the image, as well as the visual features of the images. The method states that the engagement sequence of an image can be said to depend on two independent quantities, namely scale and shape of an image. Once the shape and scale of an image have been predicted, combining them the predicted sequence of an image over 30 days is obtained. The current work follows a previous work done in the same direction, with certain speculations and suggestions of improvement. 
\end{abstract}

\section{Introduction}
In recent times, there has been a great interest in trying to find out the popularity of an image or a post. Such information holds importance for a lot of purposes. This could vary from social media campaigns, marketing, raising awareness on an issue or recommendation systems. In context of social media campaigns, the social dynamics of a post will inform a user about how long the post continues to engage the viewers. Thus, when that post becomes obsolete, another one can be posted to maximise the audience engagement.\newline
In Khosla et al. [1],the authors stated that using the social features of the user who posts the image, the social features of the image and the visual features of an image, a popularity score could be assigned to each image. Accordingly, they defined a normalised log function that would define the popularity score of an image as
\begin{equation}
score_{i} = log(c_{i}/T_{i} +1)\newline
\end{equation}
where, ${c_i}$ is the engagement score (here, the total number of views since upload)  and ${T_i}$ is the time since upload. Figure 1.1 shows the number of views of two images while Figure 1.2 shows the popularity score for two images and how it varies with time.
Previously, work related to finding out the popularity of tweets [5] and news articles had been done. However, following [1], several works followed that attempted to predict the popularity of videos and images. However, most of them predicted the cumulative engagement score as is visible in all social media. 
\begin{figure}[h]
\centering
\begin{subfigure}[h!]{0.4\linewidth}
  \includegraphics[width=\linewidth]{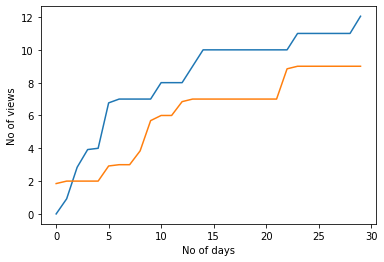}
\caption{Views of Two Images over time}
\end{subfigure}
\begin{subfigure}[h!]{0.4\linewidth}
\includegraphics[width=\linewidth]{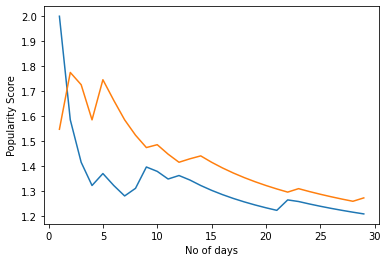}
  \caption{Popularity Score of Two Images over time}
 \end{subfigure}
 \caption{Views obtained by images and their popularity score}
\end{figure}

\section{Related Work}
Khosla et al. [1], rightly showed through their equation how the popularity of an image varied over time. A particular image remains usually popular during the initial time of upload but with time, it's popularity drops. However, the equation compared popularity of two images without considering their engagement dynamics. For instance, an image uploaded not very long ago, having a greater number of views would have a higher popularity score than another which was posted much ago but now had almost the same number of views. This was even stated in Valafar et al.[4],that the number of views increases for the first few days and then becomes stagnant slowly with time. Thus Ortis et al. [2,3] stated that the log function as suggested by [1] does not take into consideration the dynamic of the engagement of a post. This way an older photo is directly compared to a newer photo and thus punished by giving it a much lower popularity score. \newline 
In [2,3], the authors therefore stated that the engagement dynamics of an image is also important and needs to be considered while predicting the popularity of images. Thus, they defined two important parameters, the \textit{shape} and \textit{scale} of an image. They attempted to predict the number of views each image obtained over a period of 30 days.For each image, the views it receives at the end of 30 days is called the scale. Once the 30 day sequence is divided by the scale, a sequence varying between 0-1 is obtained. This sequence is called the shape. Accordingly they stated
\begin{equation}
seq_{scale}=max[ v_{1},v_{2}...v_{n} ]
\end{equation}
\begin{equation}
seq_{shape}=[v_{1}/seq_{scale},v_{2}/seq_{scale}...v_{n}/seq_{scale}]
\end{equation}
Finally predicting the scale and the sequence individually, the entire sequence is obtained as
\begin{equation}
seq=seq_{scale}*seq_{shape}    
\end{equation}

\section{Available Dataset}
Several datasets existed in prior that dealt with images and their engagement scores, many of those being Flickr images. However, most of them only contained the cumulative values of the engagement scores. Since, Ortis et al. [2] aimed to predict the number of views each image obtains over 30 days, they had created a new dataset that contained engagement scores of images for a period of 30 days. They called this dataset the \textbf{Social Image Popularity Dynamics Dataset}(SPID 2018). An extension of the previous dataset has been used for this work. This dataset was provided during the \textbf{ICIP 2020:Image Popularity Prediction Challenge.} The dataset consists of \textasciitilde{20K} Flickr images labelled with their engagement scores (here, the number of views). For each image, the dataset also includes user's and photo's social features that have been proven to have an influence on the image popularity on Flickr (e.g., number of user's contacts, number of user's groups, mean views of the user's images, photo tags, etc.). Besides these, even the image is provided so that visual features can be used to extract meaningful information from each image.
Table 3.1 shows the social features of the photos while Table 3.2 shows the social features of the users .

\section{Implemented Methodology}
In the present work, a strategy similar to that suggested in [2] has been followed. The first and foremost thing to do was data pre-processing. Since the dataset contained real world data, there were a lot of missing values. Proper data interpolation had to be done. The features that contained string were converted into embeddings. Following this, the engagement scores of the images were analysed by plotting the values. Figure 4.1(a) shows one such example.
On plotting the engagement values of different images, it was observed that two different images may have the same engagement dynamics but the views each image obtains may be very different, as was pointed out in [2]. This can be observed on comparing Figure 4.1(a) and Figure 4.1(b).

\begin{figure}[h]
\centering
\begin{subfigure}[h!]{0.4\linewidth}
  \includegraphics[width=\linewidth]{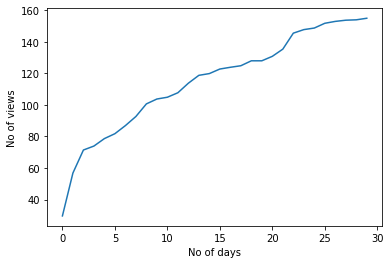}
\caption{No of Views of Image-86 over 30 days}
\end{subfigure}
\begin{subfigure}[h!]{0.4\linewidth}
\includegraphics[width=\linewidth]{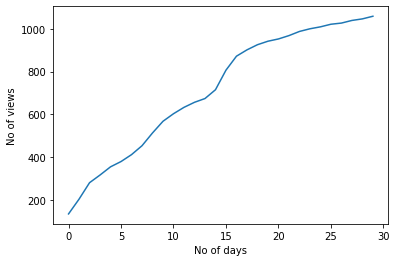}
  \caption{Actual Views of Image-13174 over 30 days}
 \end{subfigure}
 \caption{Views of two different images having similar dynamics}
\end{figure}

Therefore, this makes it difficult to compare different images because of their different scales. So to bring them all to the same scale, the engagement sequence of each image was divided by it's maximum number of views, that is the views obtained by each image till the 30th Day. The maximum views that each image obtains, is referred to as the \textit{scale} of that image. Thereby on dividing the views obtained by each image over 30 days by their respective scale, a normalised set of views is obtained for each image. Figure 4.2(a) shows an image with it's actual views while Figure 4.2(b) shows the views for the same image scaled between 0-1. 

\begin{figure}[h]
\centering
\begin{subfigure}[h!]{0.4\linewidth}
  \includegraphics[width=\linewidth]{popularity.png}
\caption{No of Views of Image-86 over 30 days}
\end{subfigure}
\begin{subfigure}[h!]{0.4\linewidth}
  \includegraphics[width=\linewidth]{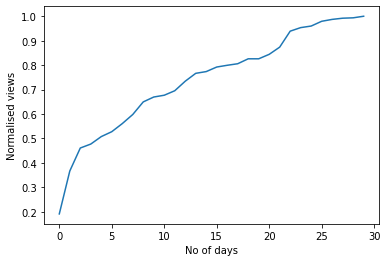}
\caption{Normalised Views of Image-86 over 30 days}
 \end{subfigure}
 \caption{Actual Views and Normalised Views of the Same Image}
\end{figure}

Once the views for all the images were scaled down to 0-1, the engagement sequence of the images along were plotted. Figure 4.3(a) shows what it looks like with the views for all the images to be scaled down to 0-1. Therefore, it can be observed that no such proper number of clusters can be visualised from it. However, on plotting 5 random images as in Figure 4.3(b), it can be seen that the shapes of some images are similar, hinting to the existence of clusters.

\begin{figure}[h!]
\centering
\begin{subfigure}[h!]{0.4\linewidth}
  \includegraphics[width=\linewidth]{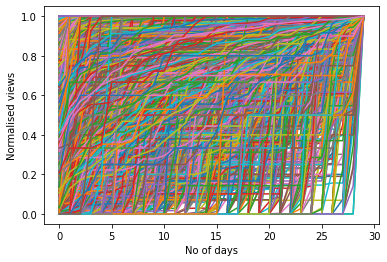}
\caption{Normalised Views of all the Images over 30 days}
\end{subfigure}
\begin{subfigure}[h!]{0.4\linewidth}
  \includegraphics[width=\linewidth]{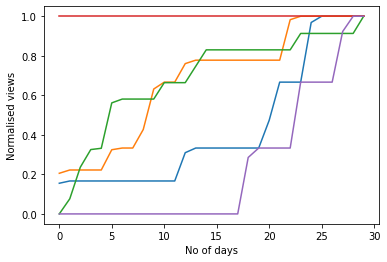}
\caption{Normalised Views of 5 Images over 30 days}
 \end{subfigure}
 \caption{}
\end{figure}
\subsection{Shape Allotment}
In [2],the authors implemented K-means for the purpose of clustering the images based on their normalised views. They observed that for \textbf{k=50}, optimal results were obtained. Therefore, in the present work, the following methods were explored to find out the optimal number of clusters that could be obtained using K-means.
\subsubsection{Elbow Method}
The dataset was clustered by varying the number of clusters from 1-80 (since the optimal suggested was 50) in an attempt to verify the optimal number of clusters. The Elbow Method shows the total WSS(Within Cluster Sum of Squares) for each cluster number. A lower WSS hints at proper clustering of the data. Once an \textit{elbow like} shape is observed, the cluster number corresponding to that point gives the optimal number of clusters. On calculating the total WSS , a graph of WSS vs No of clusters was plotted as shown in Figure 4.4.

\begin{figure}[h!]
\centering
  \includegraphics[width=\linewidth]{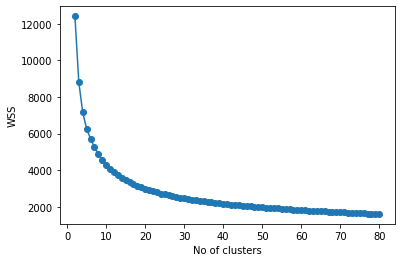}
\caption{Elbow Method Analysis}
\end{figure}

Therefore, it can be observed that no proper elbow is visible especially not at k=50 but that can be expected as Elbow Method is a Naive Approach.
 \newpage
\subsubsection{Silhouette Score}
The next method attempts to find the optimal number of k by calculating the Silhouette score for each cluster number. A Silhouette score displays a measure of how close each point in one cluster is to points in the neighboring clusters. By varying the number of clusters from 2-80 (because Silhouette score calculation demands at least 2 clusters), the Silhouette score for each cluster number was plotted as shown in Figure 4.5.
\begin{figure}[h!]
  \includegraphics[width=\linewidth]{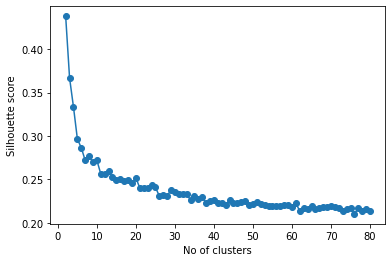}
\caption{Silhouette Score}
\end{figure}

A higher Silhouette score is desirable because it shows that the data points within a cluster are more similar to the points in the cluster and dissimilar to the points in other clusters. However, it may not always give a proper result (which is k=2 here) because it may simply imply that the clusters have been formed in such a way that the clusters are away from each other but aren't proper. Furthermore, it is obvious that k=2 cannot be possible, since from Figure 4.3(b) it can be seen that at least 3 different clusters exist. 
 \newpage
\subsection{Shape Allotment Analysis}
In \textbf{Section 4.1.1} and \textbf{Section 4.1.2}, it is observed that no optimal k was found and certainly not k=50. Furthermore, a Hierarchical clustering (Mean Shift Clustering) was also tried out to test if it performs any better than K-means. The data  was clustered  using K-means with k=50 as suggested by [2]. Whereas, with the \textit{bandwidth as 0.53} for Mean Shift clustering, \textbf{49 clusters} were obtained. A Random Forest classifier (RNF) was chosen as it uses ensemble method with several decision trees and thus gives the best of classification results. The optimal parameters for the RNF were obtained using a Grid-Search method. Thus, with the same set of parameters, a comparison between both the clustering methods was done. All the social features related to the user and posts were provided as input into the RNF.
\newline
Accuracy on using K-means Clustering = 20\%
\newline
Accuracy on using Mean Shift Clustering = 62\%
\newline
With even the number of clusters almost equal, Mean Shift clustering gave better results than K-means,thus making it our preferred choice. Once the clustering was done, the cluster centroids were treated as the \textit{shape prototypes}.Each image was then classified into their clusters using RNF. Figure 4.6 shows the cluster centroids (shape prototypes) obtained after Mean Shift Clustering.

\begin{figure}[h!]
  \includegraphics[width=\linewidth]{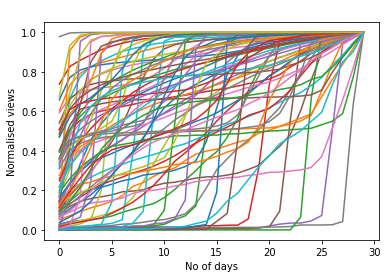}
\caption{Cluster centroids obtained by Mean Shift Clustering}
\end{figure}

\subsection{Scale Prediction}
Once the shape allotment was done, the scale of each image had to be predicted. This was done using a SVR (Support Vector Regressor). Social features related to the post were fed into the SVR one at a time. Then the Spearman's Rank Correlation of each feature with the maximum views , i.e the scale was recorded. Even the visual features were extracted from each image exploiting three state of the art Convolutional Neural Networks (CNN).(Hybridnet [6], DeepSentiBank [7] and GoogleNet [8]). Only the last two layers of activations before the softmax were used to extract the features. However, they contributed very little to predicting the scale of the image, as had been predicted in [1]. Finally, a suitable combination of the features was used so as to get a decent rank correlation. Once the necessary features were selected, the scale of the images were predicted using the SVR. Table 4.1 shows some of the highest Rank Correlations obtained by considering some individual features and groups of features.
\newline
\newline
\newline
\begin{table}[h!]
\centering
\begin{tabular}{|l|l|}
\hline
Features &
  \begin{tabular}[c]{@{}l@{}}Rank Correaltion\\     (Spearman)\end{tabular} \\ \hline
MeanViews                      & 0.7749 \\ \hline
Counts                         & 0.556  \\ \hline
GroupsCount                    & 0.562  \\ \hline
MeanViews+GroupsCount          & 0.7448 \\ \hline
MeanViews+Contacts             & 0.7465 \\ \hline
MeanViews+Contacts+GroupsCount & 0.7469 \\ \hline
\begin{tabular}[c]{@{}l@{}}MeanViews+Contacts+GroupsCount\\ +GroupsAvgMemb+GroupsAvgPic\end{tabular} &
  0.744 \\ \hline
\end{tabular}
\caption{Rank Correlation of Features}
\end{table}

\section{Results}
The training and testing of the images was done based on 9:1 train, test split. This was repeated for 10 runs and the final result was obtained by averaging the results of each run. After obtaining the scale and shape individually, on multiplying the \textit{scale} with the \textit{shape} of each image, a 30 day predicted engagement sequence of the images was obtained. However, a problem with predicting the scale was that, it was unbounded. Therefore, the model could end up predicting both very high as well as very low values. Thus, evaluating the results using an RMSE could lead to very high errors because of some outliers. Therefore, to make a more meaningful evaluation, 25\% of the highest and and the lowest predicted values were trimmed off.The final results were evaluated using the 25\% trimmed RMSE (interquartile mean) and the Median RMSE. Some of the best results obtained using different considered features and the corresponding rank correlation as well as the final errors, are presented in Table 5.1.
\newline
\newline
\newline
\begin{table}[h!]
\begin{tabular}{lllll}
\cline{1-4}
\multicolumn{1}{|l|}{Features} &
  \multicolumn{1}{l|}{RankCorrelation} &
  \multicolumn{1}{l|}{\begin{tabular}[c]{@{}l@{}}tRMSE\\ 0.25\end{tabular}} &
  \multicolumn{1}{l|}{\begin{tabular}[c]{@{}l@{}}tRMSE\\ MED\end{tabular}} &
   \\ \cline{1-4}
\multicolumn{1}{|l|}{\begin{tabular}[c]{@{}l@{}}MeanViews+Contacts+GroupsCount+\\ GroupsAvgMembers+GroupsAvgPictures+\\ NumGroups+AvgGroupsMemb+AvgGroupsPhotos\end{tabular}} &
  \multicolumn{1}{l|}{0.753} &
  \multicolumn{1}{l|}{12.2} &
  \multicolumn{1}{l|}{8.5} &
   \\ \cline{1-4}
\multicolumn{1}{|l|}{\begin{tabular}[c]{@{}l@{}}Contacts+PhotoCount+MeanViews+\\ NumGroups+AvgGroupMembers+AvgGroupPhotos\end{tabular}} &
  \multicolumn{1}{l|}{0.761} &
  \multicolumn{1}{l|}{12.06} &
  \multicolumn{1}{l|}{8.25} &
   \\ \cline{1-4}
 &
   &
   &
   &
  
\end{tabular}
\caption{Results obtained by considering some features}
\end{table}

\section{Conclusion and Improvement}
In this work, following a method similar to the one suggested in [2], the number of views an image obtains over a period of 30 days was predicted. This is much more challenging than predicting the total amount of views an image receives at the end of a period. This involves considering the engagement dynamics of an image . The shape allocation of images is something that could be worked on. Usually, when clustering is done for points in 2D , it is based on their features like an X-axis value and a Y-axis value. However, here clustering takes place based on the 30 column values. The number of columns being the coordinates on the X axis and the values of the 30 columns being the respective Y coordinates. So technically, clustering takes place using only the column values or the Y coordinates. Therefore, clustering like this isn't very representative of the shape of graph plots. It is the sequence of values that are representative of the shape of a plot and not the individual value themselves. However, clustering on the 30 columns considers the data as as individual values. Therefore, some other feature, like area under each plot could possibly be used to improve the results. The area could be a better representative of the shape and thus could lead to better clustering. \newline  Furthermore, Rank Correlation (Spearman's Rank Correlation) is sometimes considered an obsolete method of showing relation between two entities. Rank Correlation between two entities A and B has the same value as Rank Correlation between B and A. However, that need not always be true. Hence, other methods such as PPS (Predictive Power Score) could possibly be used for improving results. 
\section{References}
1.Khosla, Aditya, Atish Das Sarma, and Raffay Hamid. "What makes an image popular?." Proceedings of the 23rd international conference on World wide web.pp 867-876 ACM, 2014.
\newline2. Ortis, Alessandro, Giovanni Maria Farinella, and Sebastiano Battiato. "Prediction of Social Image Popularity Dynamics." International Conference on Image Analysis and Processing. Springer, Cham, 2019, pp 572-582.
\newline3. A. Ortis, G. M. Farinella and S. Battiato, “Predicting Social Image Popularity Dynamics at Time Zero” in IEEE Access, 2019, 7, pp. 1–15.
\newline4. Valafar, M., Rejaie, R., Willinger, W.: Beyond friendship graphs: a study of user interactions in flickr. In: Proceedings of the 2nd ACM workshop on Online social networks. pp. 25–30. ACM (2009) 
\newline5. Bandari, R., Asur, S., Huberman, B.A.: The pulse of news in social media: Forecasting popularity. ICWSM 12, 26–33 (2012)
\newline6. Zhou, B., Lapedriza, A., Xiao, J., Torralba, A., Oliva, A.: Learning deep features for scene recognition using places database. In: Advances in neural information
processing systems. pp. 487–495 (2014)
\newline7. Borth, D., Ji, R., Chen, T., Breuel, T., Chang, S.F.: Large-scale visual sentiment ontology and detectors using adjective noun pairs. In: Proceedings of the 21st ACM international conference on Multimedia. pp. 223–232. ACM (2013)
\newline8. Szegedy, C., Liu, W., Jia, Y., Sermanet, P., Reed, S., Anguelov, D., Erhan, D.,Vanhoucke, V., Rabinovich, A.: Going deeper with convolutions. In: In proceedings
of the IEEE Conference on Computer Vision and Pattern Recognition (2015)
\end{document}